\documentclass{IEEEtran}
\IEEEoverridecommandlockouts
\IEEEpubid{\begin{minipage}{\textwidth}\ \\[12pt] \centering
  Pre-print. Under review.
\end{minipage}}
\usepackage{cite}
\usepackage{amsfonts}
\usepackage{algorithmic}
\usepackage{textcomp}

\def\BibTeX{{\rm B\kern-.05em{\sc i\kern-.025em b}\kern-.08em
    T\kern-.1667em\lower.7ex\hbox{E}\kern-.125emX}}

\usepackage{tikz}
\usetikzlibrary{arrows,shapes}
\usetikzlibrary{positioning}
\usepackage{pgfplots}
\usepgfplotslibrary{statistics}
\pgfplotsset{compat=1.10}

\newlength\figureheight
\newlength\figurewidth
\usepackage{epsfig}
\usepackage{mathtools,xparse}

\usepackage{amsmath}
\usepackage{amssymb}
\usepackage{epstopdf}
\usepackage{relsize}
\usepackage{multirow}
\usepackage{float}
\usepackage{lipsum}  
\usepackage{subcaption}
\usepackage[]{algorithm2e}
\usepackage{algorithmic} 
\usepackage{mathtools}
\usepackage[keeplastbox]{flushend}

\def\arginf{\mathop{\hbox{\rm arginf }}}
\usepackage{graphicx}
\usepackage{subcaption} 
\DeclareMathOperator*{\argmin}{arg\,min} 
\DeclareMathOperator*{\argmax}{arg\,max}    

\newcommand{\twopartdef}[4]
{
	\left\{
		\begin{array}{ll}
			#1 & \mbox{with probability } #2 \\
			#3 & \mbox{with probability } #4
		\end{array}
	\right.
}    

\begin{document}

\title{Procrustes registration of two-dimensional statistical shape models without correspondences}

\author{Alma~Eguizabal,~\IEEEmembership{Student Member,~IEEE,} ~Peter~J.~Schreier,~\IEEEmembership{Senior Member,~IEEE,} and J\"urgen Schmidt.

\IEEEcompsocitemizethanks{\IEEEcompsocthanksitem A. Eguizabal and P. J. Schreier are with the Signal and System Theory Group, University of Paderborn, Germany (e-mail: alma.egui@gmail.com). J. Schmidt is with the Clinic for Trauma Surgery and Orthopaedics, Augsburg University Hospital, Germany.}       }

\maketitle

\begin{abstract}
Statistical shape models are a useful tool in image processing and computer vision. A Procrustres registration of the contours of the same shape is typically perform to align the training samples to learn the statistical shape model. A Procrustes registration between two contours with known correspondences is straightforward. However, these correspondences are not generally available. Manually placed landmarks are often used for correspondence in the design of statistical shape models. However, determining manual landmarks on the contours is time-consuming and often error-prone. One solution to simultaneously find correspondence and registration is the Iterative Closest Point (ICP) algorithm. However, ICP requires an initial position of the contours that is close to registration, and it is not robust against outliers.
We propose a new strategy, based on Dynamic Time Warping, that efficiently solves the Procrustes registration problem without correspondences. We study the registration performance in a collection of different shape data sets and show that our technique outperforms competing techniques based on the ICP approach. Our strategy is applied to an ensemble of contours of the same shape as an extension of the generalized Procrustes analysis accounting for a lack of correspondence.

\end{abstract}

\section{Introduction}
When we wish to perform statistical analysis of shape we assume that the shape observations belong to real world forms. These may come from a sketch of an archaeologist, an automatic segmentation of an image processing algorithm, or manually drawn anatomical landmarks. In this contribution we consider that a shape observation contains the vertices of an outline representing the contour of an object of interest. We consider landmark-based statistical shape models, as for instance, the Point Distribution Models (PDM) \cite{Cootes96, Eguizabal18}. Typically, when we collect shape observations to perform work with a Point Distribution Model, we assume that they all contain the same number of $N$ points, and that these points in correspondence. These points (or vertices) are also known as landmarks, since its position is generally determined by some meaningful information. For instance, in the archaeological study of spears, an evident landmark is the tip. With respect to an anatomical study of a hand outline, also the tips of the fingers are intuitive landmarks.

The training contours that are required to learn a shape model are generally manually or semi-automatically obtained from training images, which is very time-consuming. Furthermore, a Procrustes registration of the shape samples is necessary to build the statistical shape models \cite{Cootes96, Luthi13}.

Statistical shape models in segmentation problems make use of a Procrustes registration in two main scenarios.
\begin{enumerate}
	\item In the training process of the model: The training samples of the shape need to be rigidly registered (in scale, translation, and rotation) before the shape statistics are learned from the set. This is typically performed with a generalized Procrustes analysis \cite{Dryden16, Kendall84}, which is an iterative algorithm where a Procrustes registration of each training sample is performed in every iteration.
	\item During the fit of the model in a segmentation problem. The statistical shape model needs to be registered to a target contour in an image to be segmented, in order to incorporate the prior knowledge about the shape that the model provides. Examples of these segmentation algorithms are: Active Shape Models \cite{Cootes96}, Constrained Local Models \cite{Lindner13}, and deep-learning landmark localization strategies \cite{Zadeh17}.
\end{enumerate}

A Procrustes registration (rotation, scale, and translation, as defined in \cite{Dryden16}) between two landmark-based shapes with known correspondences is a linear least-squares problem: when the correspondences are available, a Procrustes analysis can be employed \cite{Gower75, Kendall84, Cootes00}. However, determining correspondences is challenging. The corresponding points are often determined manually with a few landmarks \cite{Cootes04}. The most common choice of landmarks in human-based images are anatomical references, which allow to maintain consistency between different samples in the annotation process. However, it is tedious to determine anatomical landmarks  \cite{Lindner15}. Establishing manual correspondences is done based on experience, which is not optimal. Furthermore, manual labelling is generally not very dense (the number of landmarks is much smaller than the image resolution) and the variability covered by the shape model, as well as the precision of the model, are affected. Also, the definition of an anatomical landmark typically refers to a point on the surface of the anatomy, whose projection in the image plane may not belong to the two-dimensional boundary.

In this contribution we have studied particularly an imaging technique that can benefit from PDM: fluoroscopic X-ray. This image modality is used intraoperatively and therefore computer-guided-surgery algorithms utilize it \cite{We18}. Since it is a two-dimensional image modality, when anatomical landmarks are used to register the training samples of the model, the fact that anatomy actually belongs to a three-dimensional space, is especially problematic. Also, the focal point of the image varies because the acquisition system, the C-arm, moves \cite{We18}. Therefore, another difficulty in this application arises since the anatomical contours in the images are open and cover different lengths of the boundaries.

Motivated by the challenges presented by fluoroscopic images, we have developed a strategy to perform a Procrustes registration when the correspondences are unknown (no need of manual landmarks), and the lengths of the training contours are not necessary the same (since they could have been extracted automatically).

\subsection{Related work}
When correspondences between the landmark-based shape samples are unknown, they may also be assigned automatically  with an Iterative Closest Point (ICP) approach \cite{Besl92,Hufnagel2008}. This algorithm iterates between solving the registration and finding the corresponding points, which are chosen as the closest in terms of a defined distance (e.g. Euclidean). However, the algorithm only converges to a good solution if the contours are initially close to the registration, and it is not robust against outliers. Assigning correspondences and simultaneously solving a registration problem has aroused interest before and there are other solutions, such as in \cite{Rangarajan97} and \cite{Luo03}. In \cite{Rangarajan97} the problem is presented as an extension of Procrustes alignment, and it uses a gradient descent approach to determine correspondences. In \cite{Luo03}, the authors incorporate probability and cross-entropy in the correspondence and determine a cost function to minimize. Another approach is given in \cite{Myronenko10}, which optimizes a similar objective function in a more rigorous way and presents a probabilistic approach called Coherent Point Drift (CPD). The authors in \cite{Bergstrom17} show a technique that considers M-estimators in ICP to assign the correspondences. These approaches, like ICP, typically assume that initially any point on the reference contour may correspond to any other point on the target contour. Then, the correspondence is modeled in a matching matrix whose dimensions are reference length times target length.
The proposed technique in \cite{Scott06} introduces an order-preserving constraint in the correspondences, modelling it with graphical models and solving it with dynamic programming. However, it considers only closed contours. \cite{817410} also considers a dynamic programming approach, although designed for a different application, and hence with subsequent heuristics that do not apply directly to the problems of training shape models.

There are also approaches where a group-wise registration is performed. In \cite{Bookstein97} the authors use thin-plate B-splines to model the deformation of the shapes, together with a generalized Procrustes analysis to perform the registration. However, the contours they use are not very dense (around 15 landmarks), which limits the resolution of the resulting shape model, and the thin-plate deformation may become very computationally demanding with denser contours of higher resolution. The authors in \cite{Rueckert03} define a plane-to-plane warping  based on free deformation and mutual information. They assume, however, a good initial guess in terms of rotation, scale, and translation. Similarly, in \cite{Cootes10} they also consider a plane-to-plane warping and the image appearance in the registration, but requiring a good initialization of the samples, which may be done manually. 
Alternatively, other solutions are based on a contour-to-contour warping. The approach in \cite{Glover09} performs a generalized Procrustes analysis
determining a geodesic distance between the contours. Nevertheless this distance depends on ad-hoc parameters and requires a good initialization of the registration. In \cite{Davies02} the authors describe a method to build shape models when correspondences are not available based on information theory. However, in order to establish the correspondences, they need to select a reference shape whose parameterization is fixed manually. Also, the contours are not longer than 30 points, which is orders of magnitude smaller than the typical image size.
\subsection{Contributions}
In this paper, we propose Dynamic Time Warping (DTW) \cite{Muller07} to establish an automatic correspondence between the landmark-based shapes to be registered, which avoids the need of an initial manual correspondence and same landmark-set lengths.

Hence, we extend the typical Procrustes registration considering an additional DTW step that solves the correspondences. DTW imposes constraints on the order of corresponding points between contours, leading to more accurate solutions. In our approach the input signals can contain thousands of points \cite{Rakthanmanon12}, providing higher resolution and more precision to the resulting shape models.  Our approach is also able to deal with both open and closed contours. Also, there is no need for a manual initialization of the registration.  Furthermore, in order to add robustness against outliers, our strategy combines the DTW with a weighted Procrustes registration. The resulting algorithm jointly optimizes the correspondence and the rigid registration by iterating between the DTW and the Procrustes registration steps in an alternating-optimization fashion.

We validate our approach pair- and group-wise using contours obtained from fluoroscopic images of the proximal and distal femur. The contours extracted from the images are directly the input of our registration algorithm, with no need of manual initializations. In these images, the boundaries are open and have unclear start and end points, and have also different lengths.
 
We perform a comparative study of the accuracy of our proposed Procrustes registration approach with respect to two competing techniques, \cite{Myronenko10} and \cite{Bergstrom17}, which our approach outperforms. As another example, we also consider a database of hand boundaries extracted from natural images \cite{Yoruk06}.

The paper is organized as follows. In Section II we formulate the problem, considering shape contours to be aligned with DTW. In Section III we present our solution, deriving modifications to DTW and the Procrustes registration. In Section IV we describe the experiments and the comparative studies. In Section V and VI  we discuss the results and summarize the conclusions.

\section{Preliminaries and problem formulation}
We model a contour as a complex vector, and a Procrustes registration (scale, rotation, and translation) as a complex affine transformation \cite{Dryden16}. We consider DTW to establish a high resolution correspondence between two shape vectors and formulate the registration. Then, we also propose a group-wise approach, as an extension of the generalized Procrustes analysis, to register a set of shape vectors and estimate its mean when there is no correspondence information.
 
We assume that the contours to be registered have the same shape according to Kendall's definition \cite{Kendall84}, i.e., they share the same geometrical information after scale, rotation, and translation are removed.

\subsection{Contours as complex time series}
We model the points of a shape boundary as snapshots of a complex time series whose real and imaginary parts correspond to the two coordinates of Euclidean space.  Thus, we model a collection of points as a complex vector $\mathbf{x} \in \mathbb{C}^{N\times 1} $. We assume the $N$ points in $\mathbf{x} = \big[x[1], \dots, x[N]\big]^T$ are ordered, i.e., there exists a curve topology such that the vector contains the path between the first point $x[1]$ and last point $x[N]$. The pose parameters (scale, rotation, and translation) are defined by a complex affine transformation, i.e., rigid transformation with scaling. This rigid transformation of $\mathbf{x}$ is defined as
\begin{equation}
\mathbf{x}_t = r\mathbf{x}+\mathbf{1} t , 
\end{equation} 
where $\mathbf{1}$ is a $N \times 1$ vector of ones, $\{r,t\} \in \mathbb{C }$ are the pose parameters of the transformation ($r$ is the scale in magnitude and rotation in phase, and $t$ the translation), and $\mathbf{x}_t\in \mathbb{C}^{N\times 1}$ is the transformed vector. According to Kendall's definition of shape \cite{Kendall84}, this transformation does not alter the shape of the vector, that is, $\mathbf{x}$ and $\mathbf{x}_t$ have the same shape. 

\subsection{Procrustes registration with correspondences}
\label{sec:procus}
Let us assume we have the shape vectors $\mathbf{x}_1$ and $\mathbf{x}_2$, both containing $N$ points, and that all points are in one-to-one correspondence for $n= 1,\dots,N$. Let us consider the Procrustes registration \cite{Dryden16} of $\mathbf{x}_2$ onto $\mathbf{x}_1$. This is performed by the pose parameters ${r,t}\in \mathbb{C}$ that minimize the squared distance between $\mathbf{x}_1$ and $\mathbf{x}_2$. We define this distance as
\begin{equation} \label{dp}
d^2 = \sum_{n=1}^N|x_1[n]-x_2[n]|^2=||\mathbf{x}_1 - \mathbf{x}_2||^2,
\end{equation}
where $|\cdot|$ denotes absolute value, and $||\cdot||$ $l_2$-norm (or Euclidean distance). Let us rewrite the transformation as $r\mathbf{ x}_2+\mathbf{1}t=[\mathbf{ x}_2\quad \mathbf{1}]\mathbf{p}$,
where $\mathbf{p} = [r \quad t]^T$ is a vector containing the pose parameters. The vector $\mathbf{p}$ that minimizes the squared distance between $\mathbf{x}_1$ and $r\mathbf{x}_2 + \mathbf{1}t$ is the solution to a linear least-squares fit. Let us define the matrix $\mathbf{X}_2=[\mathbf{ x}_2\quad \mathbf{1}]$. The pose $\mathbf{p}^\star$ that minimizes the distance after the Procrustes registration is 
\begin{equation} \label{procrus}
\mathbf{p}^\star= \arg\min_{r,t}|| \mathbf{x}_1 - (r\mathbf{x}_2+\mathbf{1} t) ||^2,
\end{equation}
 where $\mathbf{p}^\star= (\mathbf{ X}_2^H\mathbf{ X}_2)^{-1}\mathbf{ X}_2^H\mathbf{x}_1$.

\subsection{Dynamic Time Warping to establish correspondences}
A Procrustes registration as defined in \eqref{dp} requires a one-to-one correspondence between vector elements. In most applications this is not available. In order to assign correspondences between two shape vectors, we consider DTW. This technique is commonly used in time series analysis to find an optimal alignment between two signals \cite{Muller07}. A non-linear warping of the signals is considered in order to determine the corresponding points. The objective of DTW is to align these signals so that the sum of the distances (e.g. Euclidean) between the corresponding points is smallest. 

Let us assume that $\mathbf{x}_1$ and $\mathbf{x}_2$, containing $N_1$ and $N_2$ points respectively, are two shape vectors with unknown correspondences. We would like to determine the optimal warping-path matrix $\mathbf{C}\in\mathbb{R}^{2\times L}$ in terms of Euclidean distance. This matrix $\mathbf{C}$ is composed of $L$ correspondence vectors $\mathbf{c}_l = (n_1^{(l)},n_2^{(l)})^T$, with $l = 1, \dots, L$. The vectors $\mathbf{c}_l$ establish correspondences between points $x_1[n_1^{(l)}]$ and $x_2[n_2^{(l)}]$. The optimal warping path $\mathbf{C}^\star$ between $\mathbf{x}_1$ and $\mathbf{x}_2$ is the one that minimizes the sum of distances
\begin{equation} \label{DTW}
\mathbf{C}^\star=\arg\min_{\mathbf{C\in \mathbb{P}}} \sum_{l=1}^{L}|x_1[n_1^{(l)}]-x_2[n_2^{(l)}]|^2 ,
\end{equation}
 where $\mathbb{P}$ is the set of allowed warping paths. A warping path in $\mathbb{P}$ must satisfy the following contraints \cite{Muller07}: 

\begin{enumerate}
\item Boundary condition: The beginning and end points of the shape vectors $\mathbf{x}_1$ and $\mathbf{x}_2$ are in correspondence, i.e., $\mathbf{c}_1 = (1,1)$ and $\mathbf{c}_L = (N_1,N_2)$.
\item Monotonicity condition: The topology of the curve is respected in the correspondence assignment, meaning that $n_i^{(1)} \leq n_i^{(2)} \leq \dots  \leq n_i^{(L)}$ for $i = 1,2$.
\item Step size condition: Each element of $\mathbf{x}_1$ corresponds to at least one element in $\mathbf{x}_2$ and vice versa. Therefore, the elements in the matrix $\mathbf{C}$ satisfy $\mathbf{c}_{l+1}-\mathbf{c}_{l} \in \{(1,0),(0,1),(1,1)\}$ for all $l \in \{1, \dots, L-1\}$.

\end{enumerate}

These constraints on the warping path allow an efficient computation of $\mathbf{C}^\star$ with dynamic programming \cite{Muller07}. We define the resulting shape vectors in point-to-point correspondence as $\tilde{\mathbf{x}}_1 = \big[x_1[n_1^{(1)}],\dots,x_1[n_1^{(L)}]\big]^T$ and $\tilde{\mathbf{x}}_2 = \big[x_2[n_2^{(1)}],\dots,x_2[n_2^{(L)}]\big]^T$. Each vector contains $L$ points, and $\max(N_1,N_2) \leq L \leq N_1+N_2-1$. 

To sum up, DTW determines a warping path between two shape vectors that is optimal in terms of the sum of Euclidean distances between corresponding points. Considering the path constraints, there is at least one corresponding point in $x_1[n]$ and $x_2[n]$ for every $n$, but there may also be more than one point in $x_1[n]$ corresponding to one point in $x_2[n]$ and vice versa. We call this a many-to-one correspondence. 

\subsection{Simultaneously determining correspondences and registration}
When we employ DTW to find a path of correspondence between two boundaries, the result may change when a linear transformation is applied to the boundaries. The correspondence path that best explains the shape deformation is the one considered after a Procrustes registration between the boundaries. 
At the same time, this Procrustes registration depends on the determined correspondences. Hence, we need to obtain the parameters that solve the overall minimization problem
\begin{equation} \label{DTWPose}
[r^\star, t^\star, \mathbf{C}^\star]=\arg\min_{r, t, \mathbf{C\in \mathbb{P}}} \sum_{l=1}^L|x_1[n_{1}^{(l)}]-(rx_2[n_{2}^{(l)}]+t)|^2 ,
\end{equation}
where the transformation parameters $r,t$ and warping path $\mathbf{C}$ are mutually dependent. 

\subsection{Group-wise correspondence and registration} \label{sec:gwr}
In order to learn statistical shape models we need to perform the group-wise registration of a set of training shape vectors \cite{Cootes96}, \cite{Dryden16}. Let us consider a set of $M$ shape vectors of the same length $N$ and with one-to-one correspondence, i.e., each $n$th element of each of the $M$ vectors corresponds to the same landmark.
Recalling the definition of shape in \cite{Kendall84}, shape variability is what remains after accounting for scale, translation and rotation (i.e. after a group-wise Procrustes registration). A typical way to remove the effects of size and translation is to normalize the shape vectors to unit size and translate their centroid to the origin of coordinates \cite{Kendall89}. That is, for a given shape vector $\mathbf{x}$, we obtain
\begin{equation} \label{preshape}
\mathbf{x}_o = \mathbf{x}-\frac{1}{N}\sum_{n=1}^Nx[n], \: \: \text{and} \: \:
\boldsymbol{\tau} = \frac{\mathbf{x}_o}{||\mathbf{x}_o||},
\end{equation}
where $\mathbf{x}_o$ is moved to the origin, and dividing by ${||\mathbf{x}_o||}$ is a size normalization. We call the vector $\boldsymbol{\tau}$ a preshape, equivalently to the geometrical definition in\cite{Kendall89}. The term preshape refers to the fact that the vector is one step away from registration since rotation still needs to be removed \cite{Dryden16}. Thus, the group-wise registration problem is reduced to find rotations.

The preshapes belong to a hypersphere, where the distance between two preshapes is a geodesic \cite{Dryden16} defined as $d_s(\boldsymbol{\tau}_1 , \boldsymbol{\tau}_2 ) = \cos^{-1}|{\boldsymbol{\tau}}_2^H{\boldsymbol{\tau}}_1 |$.
Considering we have a set of $M$ training shape vectors, and their preshapes are $\boldsymbol{\tau}_1, \ldots, \boldsymbol{\tau}_M$, the mean shape is defined as
\begin{equation} \label{meanRi}
\boldsymbol{\mu}^\star = \arginf_{\boldsymbol{\mu}} \sum_{m=1}^M d_s(\boldsymbol{\tau}_m, \boldsymbol{\mu}),
\end{equation}
where $\boldsymbol{\mu}$ is also considered a preshape so that $||\boldsymbol{\mu}||^2 = 1$ and $\frac{1}{N}\sum_{n=1}^N \mu [n] = 0$. The solution to \eqref{meanRi} is typically obtained through a generalized Procrustes analysis \cite{Gower75}, \cite{Dryden16}. This analysis iterates between minimizing the distances of the preshapes to $\boldsymbol{\mu}$ and estimating $\boldsymbol{\mu}$. 
However, when we assume not to have correspondences and, furthermore, the observed vectors may be occluded at the extremes, it is not possible to determine the preshape space and thus, compute the mean shape as in \eqref{meanRi}. 
 
\subsection{Statistical Shape Models}
In a statistical linear model of shape \cite{Luthi18}, an observed shape vector $\mathbf{x}_m$ follows 
\begin{equation} \label{shapemodel}
\mathbf{x}_m = r_m( \boldsymbol{\mu} + \boldsymbol{\delta}_m) +\mathbf{1} t_m, 
\end{equation}
where $r_m$ and $t_m$ are the Procrustes registration parameters to minimize the squared distance between $\mathbf{x}_m$ and the mean shape $\boldsymbol{\mu}$, and $\boldsymbol{\delta}_m$ is a realization of the random vector that models shape variability, typically Gaussian $\boldsymbol{\delta}_m \sim \mathcal{CN}(\mathbf{0},\boldsymbol{\Sigma})$ \cite{Goodall91}.

The parameters of a shape model, $ \boldsymbol{\mu}$ and $\boldsymbol{\Sigma}$ are learned from a set of training samples after these have been group-wise registered.
\section{DTW-based solution}
Our goal is to determine a Procrustes registration of shape vectors when the correspondences are not available. Our approach proceeds along the following lines. We first use DTW to compute a dense correspondence between two shape vectors and thus perform a Procrustes registration. However, the distance in \eqref{dp} is not robust against outliers, that is, spurious points of the vector that do not belong to the actual contour. These may occur, for example, when an edge detector selects parts of a neighboring object.

Hence, we propose a probabilistic interpretation of the Procrustes registration in \eqref{dp} to add robustness against the outliers in the assigned correspondences.

We then solve an overall minimization problem to determine simultaneously the registration and the correspondence allocation. In order to overcome the dependencies between registration and correspondence parameters, while aiming for a tractable solution, we propose an alternating optimization, where we find independently a solution for $\mathbf{p}$ (registration) and $\mathbf{C}$ (warping path of correspondence).

Furthermore, we formulate a group-wise registration approach, in which we calculate the mean shape and hence extend the generalized Procrustes analysis to deal with the lack of correspondence.
 
 
\subsection{A probabilistic Procrustes registration}
Let us recall the vectors $\mathbf{x}_1$ and $\mathbf{x}_2$ with no assigned correspondences. After DTW we obtain the vectors  $\tilde{\mathbf{x}}_1$ and $\tilde{\mathbf{x}}_2$, which are in one-to-one correspondence and have the same length $L$. Let us define the distance vector $\tilde{\mathbf{d}} = \tilde{\mathbf{x}}_1-(r\tilde{\mathbf{x}}_2+\mathbf{1} t)$, where $r$ and $t$ are the pose parameters of the registration we need to obtain. Let $\tilde{d}_l$ denote the elements in the vector $\tilde{\mathbf{d}} = [\tilde{d}_1,\ldots,\tilde{d}_L]^T$. 
Thus, $\tilde{d}_l=\tilde{{x}}_1[l]-(r\tilde{{x}}_2[l]+t)$ is the distance between the $l$th corresponding point, whose contribution should be considered in the registration calculation only if there is true correspondence between $\tilde{{x}}_1[l]$ and $\tilde{{x}}_2[l]$. Therefore, we give a probabilistic interpretation to $\tilde{d}_l$.

We model the probability of correspondence as a Bernoulli random variable $\gamma_l$ that takes value $1$ when there is correspondence, so
\begin{equation*}
\gamma_l = \twopartdef { 1 } {w_l \text{ (correspondence)}} {0} {1-w_l \text{ (no correspondence).}}
\end{equation*}
Therefore, the registration distance to consider is $\tilde{{d}_l}\gamma_l $, which is random. Instead of minimizing the sum of squared distances, as in \eqref{procrus}, the proposed problem is to minimize the expected value of the sum of these squared random distances, that is
\begin{equation} \label{E}
\mathbb{E} \left[\sum_{l=1}^{L} \tilde{d_l}^2  \gamma_l \right]= \sum_{l=1}^{L} \tilde{d_l}^2 \mathbb{E} [ \gamma_l ]=  \sum_{i=1}^{L} \tilde{d_l}^2w_l = \tilde{\mathbf{d}}^H\mathbf{W}\tilde{\mathbf{d}}  ,
\end{equation}
which results in a weighted least-squares formulation, where $\mathbf{W}$ is a diagonal matrix whose diagonal elements are the probabilities $w_1, w_2, \dots, w_L$, which will be determined below.

The pose vector $\mathbf{p}^\star=[r^\star \quad t^\star]^T$ that minimizes the sum of probabilistic distances defined in \eqref{E} is the solution to the weighted least-squares problem
\begin{equation} \label{WLS}
\mathbf{p}^\star= \arg\min_{\mathbf{p}}\tilde{\mathbf{d}}^H\mathbf{W}\tilde{\mathbf{d}} \: ,
\end{equation}
which is $\mathbf{p}^\star= (\tilde{\mathbf{ X}}_2^H\mathbf{W}\tilde{\mathbf{ X}}_2)^{-1}\tilde{\mathbf{ X}}_2^H\mathbf{W}\tilde{\mathbf{x}}_1$ with $\tilde{\mathbf{ X}}_2=[\tilde{\mathbf{ x}}_2\quad \mathbf{1}]$. 
\subsection{Determining the weights} \label{sec:DetW}
In order to obtain the correspondence probabilities, i.e., the weights $w_l$, we first transform $\tilde{\mathbf{x}}_1$ and $\tilde{\mathbf{x}}_2$ into their respective preshapes $\tilde{\boldsymbol{\tau}}_1$ and $\tilde{\boldsymbol{\tau}}_2$, as defined in \eqref{preshape}. To this end, we assume that the shape vectors $\tilde{\mathbf{x}}_1$ and $\tilde{\mathbf{x}}_2$ belong to the same statistical shape model as defined in \eqref{shapemodel}. This model is determined by the mean $\boldsymbol{\mu}$ and a deformation random vector $\boldsymbol{\delta}$. Assuming there is no prior information about the shape model, we fix the preshape of $\tilde{\mathbf{x}}_1$, i.e., $\tilde{\boldsymbol{\tau}}_1$, to be the mean. Then, the preshape of $\tilde{\mathbf{x}}_2$, i.e., $\tilde{\boldsymbol{\tau}}_2$, can be expressed in terms of $\tilde{\boldsymbol{\tau}}_1$ as
\begin{equation}
\tilde{r} \tilde{\boldsymbol{\tau}}_2 = \tilde{\boldsymbol{\tau}}_1 + \boldsymbol{\delta}',
\end{equation}
where $\tilde{r} = \exp(-j \text{arg}(\tilde{\boldsymbol{\tau}}_1^H\tilde{\boldsymbol{\tau}}_2))$, and $\boldsymbol{\delta}' = [\delta_1', \dots, \delta_L']^T$ is an obervation of the random vector of shape deformation. Gaussian deformation models are typically used in shape analysis to model non-rigid deformations \cite{Luthi18}. Hence, we choose $\boldsymbol{\delta}' \sim \mathcal{CN}(\mathbf{0}, \sigma^2\mathbf{I})$, where $\mathbf{I}$ is the identity matrix. Therefore, each $\delta_l'$ is assumed to be independent and identically distributed (i.i.d.) as complex Gaussian with mean zero and variance $\sigma^2$.
Real and imaginary parts are independent, each with variance $\frac{\sigma^2}{2}$ . We may thus employ a Chi-squared test since 
the normalized distances $\Delta_l$, for $l=1,\dots,L$, are Chi-squared distributed with two degrees of freedom, i.e.,
\begin{equation}
 \Delta_l = \frac{2|\delta_l'|^2}{\sigma^2} \sim {\chi}^2(2).
\end{equation}
Therefore, the weights $w_l$ may be determined from the cumulative  Chi-squared distribution function as
\begin{equation} \label{weights}
w_l = 1- \int_{0}^{\Delta_l}\frac{e^{-u/2}}{2\Gamma(1)} du.
\end{equation}
Since the elements of $\boldsymbol{\delta}'$ are i.i.d., we estimate $\sigma^2$ as the sample variance, that is, $\hat{\sigma}^2 = \frac{1}{L}(\tilde{r}\tilde{\boldsymbol{\tau}}_2-\tilde{\boldsymbol{\tau}}_1)^H(\tilde{r}\tilde{\boldsymbol{\tau}}_2-\tilde{\boldsymbol{\tau}}_1)$.

\begin{figure}[t] 
\centering
\includegraphics[width=.4\textwidth]{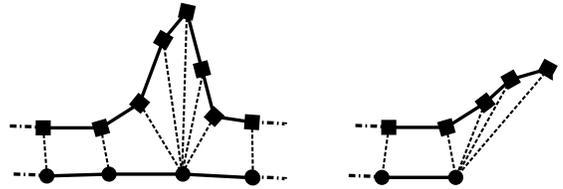}
\caption{Two examples of arc-length fluctuations. Left: There is a many-to-one correspondence due to local differences in the arc-length. Right: The squares-line is longer than the circles-line at the end, and thus many points from the squares-line correspond to a single point from the circles-line.}
\label{fig:many2one}
\end{figure}

\begin{figure}[t]
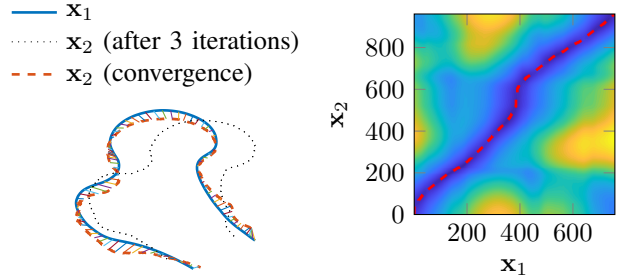

\begin{minipage}{.22\textwidth}
\centering
\input{shapes_new2}
\end{minipage}
\begin{minipage}{.22\textwidth}
\centering
\input{NewNewCost2}
\end{minipage}
\caption{Left: Result from Algorithm 2 after 3 iterations (black, dotted) and converged result (red, dashed). Right: DTW cost function for every pair of points, and corresponding warping path of correspondence between the two boundaries after convergence. More results of the working procedure are displayed in the MP4 video clip in the Supplementary Material.}
\label{fig:Exshapes2}

\end{figure}

\subsection{Soft boundary condition}
One of the constraints in DTW is the boundary condition, which requires that the beginning and end points of the two contours are always in correspondence. Therefore, when dealing with contours of different lengths, the DTW algorithm assigns the same correspondences to the additional points at the extremes, as illustrated in Fig. \ref{fig:many2one} (right), generating many-to-one correspondences at the start or end of the warping path. These many-to-one correspondences represent a missing part of the contour and should not influence the weighted least-squares minimization in \eqref{WLS}. We detect these situations by analyzing the corresponding values in the warping path matrix $\mathbf{C}$ at the start ($l=1$) and end ($l=L$). Then, we assign zero weight to any repeated corresponding points. This can be achieved with Algorithm \ref{alg:weigths}.
\begin{algorithm}
\textbf{Inputs:} $\mathbf{W}$ with diagonal elements $w_l = 1 - \int_{0}^{\Delta_l}\frac{e^{-u/2}}{2\Gamma(1)} du \quad \forall  l = 1,\dots,L$ (as in \eqref{weights}). \\ 
Path vectors $\mathbf{c}_l=(n_1^{(l)},n_2^{(l)})\quad \forall  l = 1,\dots,L$. \\
 \For{$i = \{1,2\}$}{
 	\begin{enumerate}

 	 \item $l' =\argmax_l ( n_{i}^{(1)} = n_{i}^{(l)})$,  
 	$w_1,\dots,w_{l'-1} = 0$ \\
 	\item $l'' =\argmin_l (n_{i}^{(L)} = n_{i}^{(l)})$, 
 	$w_{l''+1},\dots,w_{L} = 0$.
 	 	
 	\end{enumerate}
 }
 \caption{Adjusting the weights to account for many-to-one correspondences in $\mathbf{W}$ that are due to the soft boundary condition.}\label{alg:weigths}
\end{algorithm}

\subsection{Simultaneous pose and correspondences estimation}
The whole procedure of jointly estimating the Procrustes registration and correspondences works as follows. We solve the minimization problem
\begin{equation} \label{DTWSolution}
[r^\star, t^\star, \mathbf{C}^\star]=\arg\min_{r, t, \mathbf{C\in \mathbb{P}}} (\tilde{\mathbf{x}}_1 - \tilde{\mathbf{x}}_2)^H\mathbf{W}(\tilde{\mathbf{x}}_1 - \tilde{\mathbf{x}}_2)
\end{equation}
iteratively with an alternating optimization. The solution is described in detail in Algorithm \ref{alg:alt1}. In Fig. \ref{fig:Exshapes2} we show one example of the warping path $\mathbf{C}$ as well as the resulting cost function and registration obtained by the proposed algorithm. 

\begin{algorithm}
 \textbf{Input:} $\mathbf{x}_1 \in \mathbb{C}^{N_1}$ (reference) and $\mathbf{x}_2\in \mathbb{C}^{N_2}$ (target). \\
 \KwResult{Correspondence $\mathbf{C}$ and pose $\{r,t\}$.}
 initialization:\: $i = 0$, $c>c_{\operatorname{min}}$, $\mathbf{y}^{(0)} = \mathbf{x}_2$ \\
 \While{$c>c_{\operatorname{min}} \quad \text{and} \quad i <i_{\operatorname{max}}$}{
 \begin{enumerate}
 	\item  Find correspondence $\mathbf{C}$ between $\mathbf{y}^{(i)}$ and $\mathbf{x}_1$, as defined in \eqref{DTW}.
 	\item Use $\mathbf{C}$ to determine $\tilde{\mathbf{y}}$ and $\tilde{\mathbf{x}}_1$. 
 	\item Calculate $\mathbf{W}$ as in Algorithm \ref{alg:weigths}.
 	\item Find pose parameters $\mathbf{p} = [r \quad t]^T$ as in \eqref{WLS}. 
 	\item Determine $\mathbf{y}^{(i+1)} = r\mathbf{y}^{(i)} +\mathbf{1} t$.
 	\item $c = ||\mathbf{y}^{(i)}-\mathbf{y}^{(i+1)}||^2$, $i=i+1$
 \end{enumerate}
 }
 \caption{Proposed algorithm to determine simultaneously the registration and correspondences based on dynamic time warping.}\label{alg:alt1}
\end{algorithm}

\begin{figure*}[t]
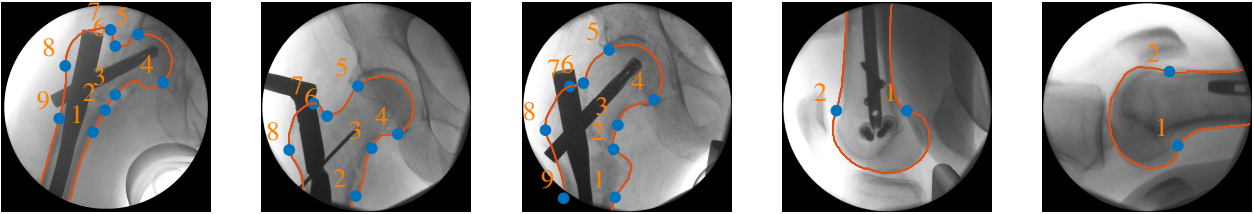

\input{head1}
\input{head2}
\input{head3}
\input{cond1}
\input{Cond_nex_ex}
\caption{Fluoroscopic images of the femur in our database. The proximal femur (first three images) contains 9 manually annotated landmarks: the first and second images have different lengths due to movements of the C-arm; the third image differs due to a more medio-lateral position. The distal femur (last two images) contains two manual landmarks, and the lengths of the visible shaft may be different.}\label{fig:Carm}
\end{figure*}

\subsection{Group-wise solution} 
Within a set of $M$ shape vectors with no assigned correspondences, we aim at minimizing the DTW distance between their mean $\boldsymbol{\mu}$ and their pose- and correspondence-corrected versions and hence extending the generalized Procrustes analysis in \cite{Dryden16}. Starting from the proposed pair-wise solution, let us assume that $\mathbf{x}_1$ is a reference vector and $\mathbf{x}_2$ is a target vector and determine the corresponding points with respect to the reference vector. We define $\hat{\mathbf{x}}_2(\mathbf{x}_1)$, of length $N_1$, to contain those points of the target $\mathbf{x}_2$ that correspond to the points in the reference $\mathbf{x}_1$. Since $\mathbf{x}_1$ and $\mathbf{x}_2$ do not have to have the same lengths and due to possible many-to-one correspondences, we agree on the following:

\begin{itemize}
\item If there is a many-to-one corresponding set of points such as in Fig. \ref{fig:many2one} (left), and there are more points in the target than in the reference, we keep only the point that is closest to the mid-point of the fluctuation.
\item If the many-to-one correspondences occur the opposite way (more points in the reference than in the target) we assume the points are repeated, unless they appear at the extremes (as in Fig. \ref{fig:many2one}, right). In such a case, in order to force the length of $\hat{\mathbf{x}}_2(\mathbf{x}_1)$ to be $N_1$, these points are labeled as ``empty".
\item If the target is longer at the extremes than the reference, we delete the extra points in the target.
\end{itemize}
We need to register a set of shape vectors to its estimated mean $\boldsymbol{\mu}$. Following the above agreements, for each of the $M$ training vectors  $\mathbf{x}_m$, $m=1,\dots,M$, we determine the vector $\hat{\mathbf{y}}_m(\boldsymbol{\mu}) = r_m\hat{\mathbf{x}}_m(\boldsymbol{\mu}) +\mathbf{1} t_m$, which is registered to the mean $\boldsymbol{\mu}$ with one-to-one correspondence. Then, the mean shape is estimated as
\begin{equation} \label{mean}
\boldsymbol{\mu}^\star = \arg\min_{\boldsymbol{\mu}} \sum_{l=1}^M||\boldsymbol{\mu}-\hat{\mathbf{y}}_m(\boldsymbol{\mu})||^2.
\end{equation} 
Notice that $\boldsymbol{\mu}$ is a preshape.
In order to register each $\tilde{\mathbf{x}}_m$ to the mean $\boldsymbol{\mu}$ we need to determine 
\begin{equation} \label{overall}
[r^\star_m, t^\star_m, \mathbf{C}^\star_m]=\arg\min_{r, t, \mathbf{C}\in\mathbb{P}} (\boldsymbol{\mu} - \hat{\mathbf{y}}_m(\boldsymbol{\mu}))^H\mathbf{W}_m(\boldsymbol{\mu} - \hat{\mathbf{y}}_m(\boldsymbol{\mu})).
\end{equation}
The expression in \eqref{overall} is solved for all $m=1,\dots,M$. As expected, the estimation of the mean in \eqref{mean} depends on the extraction of the registration parameters in \eqref{overall}. This is handled in an iterative process. 
We describe this method in detail in Algorithm \ref{alg:alt2}. Notice that the points that are labeled as ``empty" are not used when computing the mean. 

\begin{algorithm}
 \textbf{Input:} $\mathbf{x}_m$, for $m= 1, \dots, M$, with lengths $N_m$ and no assigned correspondences.\\
 \KwResult{Correspondence $\mathbf{C}_m$, poses $\{r_m,t_m\}$ and mean $\boldsymbol{\mu}$.}
 initialization:\: $i =0$, $c>c_{\operatorname{min}}$ \\
 $\boldsymbol{\mu}^{(0)}=\boldsymbol{\tau}_{m}$ (preshape of any $\mathbf{x}_m$, preferably the longest)\\
 \While{$c>c_{\operatorname{min}} \quad \text{and} \quad i <i_{\operatorname{max}}$}{
  \For{$m= 1, \ldots, M$}{
 \begin{enumerate}
 	\item  Use Algorithm \ref{alg:alt1} to find $\mathbf{C}_m$, $r_m$ and $t_m$ \\(inputs: $\boldsymbol{\mu}^{(i)}$ and $\mathbf{x}_m$).
 	\item Determine $\mathbf{y}_m = r_m\mathbf{x}_m + \mathbf{1}t_m$.
 	\item Calculate $\hat{\mathbf{y}}_m(\boldsymbol{\mu})$ using $\mathbf{C}_m$.
 \item 	$\mathbf{x}_m = \mathbf{y}_m$.
 \end{enumerate} 

}

 \begin{enumerate}
\item Compute $\boldsymbol{\mu}^{(i+1)}$ with elements 
$\mu^{(i+1)}[n] = \frac{1}{M_n}\sum_{m \in \mathcal{M}_n}\hat{y}_m(\boldsymbol{\mu})[n]$ 
\\where $\mathcal{M}_n$ is the set of indices such that $\hat{\mathbf{y}}_m(\boldsymbol{\mu})$ \\is not empty, and $M_n$ the number of such indices.
\item $c = ||\boldsymbol{\mu}^{(i+1)}-\boldsymbol{\mu}^{(i)}||^2$, $i=i+1$
 \end{enumerate} 
 }

 \caption{Group-wise correspondence and registration as an extended generalized Procrustes analysis.}
 \label{alg:alt2}
\end{algorithm}

\section{Materials and Experiments}
We validate our proposed strategy with contours of the femur extracted from fluoroscopic (low dose, and thus low quality) X-ray images. In this context, a registration technique may be used to design a statistical shape model of the femur for automatic segmentation \cite{Eguizabal17}. Our database contains manually traced contours of both proximal and distal sides of the femur. 
We also apply our technique to an open source database of hand images \cite{Yoruk06}, where the contours are automatically extracted with conventional edge detectors.

\subsection{Materials}
\subsubsection{Proximal femur}
This set contains contours from the proximal femur extracted from fluoroscopic X-rays images. These images are in anterior-posterior orientation and acquired with a C-arm during surgery implanting a cephalomedullary nail for osteosynthesis, which treats fractures of the femur\footnote{The images were routinely acquired during surgery at the Clinic for Trauma Surgery and Orthopaedics at Augsburg University Hospital. All applicable data privacy regulations were observed, and we only worked with anonymized data.}. Our X-ray images therefore also show implants and surgical tools (nail, blade, or k-wire), as seen in Fig. \ref{fig:Carm}. We collected 350 manually drawn boundaries that contain between 500 and 1000 points, are one pixel wide, and 8-connected. On those contours, 9 landmarks are manually annotated. We agreed on the following definitions of landmarks: (1) beginning of lesser trochanter, (2) end of lesser trochanter, (3) beginning of femoral neck, (4) intersection neck / femoral sphere (distal), (5) intersection neck / femoral sphere (proximal), (6) end of femoral neck, (7) superior border of greater trochanter, (8) inferior end of greater trochanter, (9) lateral point such that the connecting line to point 1 is perpendicular to the shaft axis. We show some examples in Fig. \ref{fig:Carm}.
\subsubsection{Distal femur} This set contains contours from the distal part of the femur, also extracted from fluoroscopic images from the same surgical interventions as with the proximal femur \cite{We18}. The images show the femur contour and the medial condyle. We collected 116 manually drawn contours and on those manually placed two landmarks that delimit the condyle. The contours contain between 450 and 950 points and are one pixel wide and 8-connected.
We show some examples in Fig. \ref{fig:Carm}.
\subsubsection{Hands} The set contains 1000 closed-form contours that are automatically extracted with conventional edge detectors from natural images of hands. The starting point of the contour is automatically obtained as approximately located around the wrist \cite{Yoruk06}. The contours contain between 1200 and 2200 points. This data set does not contain manually annotated landmarks.
\subsection{Competing techniques} \label{sec:CompStr}
We present an overview of the competing strategies that we consider in our experiments.
\subsubsection{Manual correspondences}
We consider a Procrustes registration based on manually determined correspondences. This is the typical registration to train statistical shape models \cite{Lindner13}. These correspondences are determined using the available manually annotated landmarks on the femur. In order to account for more corresponding points, a fixed number of equidistant landmarks is additionally extracted between the manual landmarks \cite{Cootes04}. This process is time-consuming and error-prone because the manual landmarks were tedious to annotate and difficult to define. Due to the C-arm movements, ensuring consistency of the annotated landmarks is very challenging. Also, anatomically meaningful points, such as ``the most proximal point to greater trochanter", are difficult to determine in the image or may not even lie on the bone boundary. 
With manual correspondences the registration parameters are obtained by a least-squares minimization as defined in \eqref{procrus}.

\subsubsection{Rigid Coherent Point Drift \cite{Myronenko10}} This iterative algorithm, based on ICP, models the points on the contours as Gaussian mixtures in order to add robustness to the registration. In each iteration, a correspondence matrix that accounts for every possible pair of correspondences is computed and used to determine the registration parameters. We have considered the rigid-registration version, where the obtained registration parameters are scale, rotation and translation.

\subsubsection{Robust Iterative Closest Point \cite{Bergstrom17}}
This is an iterative algorithm based on ICP. It enhances the robustness of ICP through M-estimators. The correspondences are calculated in a way similar to the CPD strategy.

\begin{figure}[t]
\centering
\begin{subfigure}[b]{0.23\textwidth}
%
%
\definecolor{mycolor4}{rgb}{0.00000,0.44700,0.74100}%
\definecolor{mycolor1}{rgb}{0,0,0}%
\definecolor{mycolor2}{rgb}{0.85000,0.32500,0.09800}%
\definecolor{mycolor3}{rgb}{0.92900,0.69400,0.12500}%
\begin{tikzpicture}

\begin{axis}[%
width=0.23\figurewidth,
height=0.23\figurewidth,
at={(0in,0in)},
scale only axis,
xmin=100,
xmax=400,
ymin=200,
ymax=500,
hide x axis,
hide y axis,
x dir=reverse,
y dir=reverse,
axis background/.style={fill=white},
legend style={legend cell align=left, at={(0.5,1.03)},legend columns=2, column sep=1, fill=none, align=left, anchor=south, draw=none, font=9}
]

\addplot  [color=mycolor1, draw=none, mark=*, mark options={solid, mycolor1}, mark size=.5pt]
  table[row sep=crcr]{%
284	471\\
284	470\\
284	469\\
284	468\\
283	467\\
283	466\\
283	465\\
283	464\\
282	463\\
282	462\\
282	461\\
281	460\\
281	459\\
280	458\\
280	457\\
280	456\\
280	455\\
279	454\\
279	453\\
279	452\\
278	451\\
277	450\\
277	449\\
276	448\\
276	447\\
275	446\\
275	445\\
275	444\\
274	443\\
274	442\\
273	441\\
273	440\\
272	439\\
271	438\\
271	437\\
271	436\\
271	435\\
270	434\\
270	433\\
270	432\\
270	431\\
270	430\\
269	429\\
269	428\\
268	427\\
267	426\\
266	425\\
266	424\\
266	423\\
265	422\\
264	421\\
264	420\\
264	419\\
263	418\\
264	418\\
264	417\\
264	416\\
263	415\\
263	414\\
263	413\\
262	412\\
262	411\\
261	410\\
260	409\\
260	408\\
260	407\\
259	406\\
259	405\\
258	404\\
257	403\\
256	402\\
256	401\\
255	400\\
254	400\\
253	400\\
252	399\\
252	398\\
252	397\\
251	396\\
250	396\\
249	395\\
248	394\\
247	393\\
247	392\\
247	391\\
246	390\\
245	390\\
244	389\\
243	388\\
242	387\\
242	386\\
242	385\\
241	384\\
240	383\\
239	382\\
238	381\\
237	380\\
237	379\\
236	378\\
236	377\\
236	376\\
235	375\\
234	374\\
234	373\\
234	372\\
234	371\\
233	370\\
233	369\\
233	368\\
232	367\\
231	366\\
230	365\\
230	364\\
230	363\\
229	362\\
228	361\\
227	360\\
227	359\\
227	358\\
226	357\\
225	356\\
224	355\\
223	354\\
223	353\\
222	352\\
221	351\\
221	350\\
220	349\\
219	348\\
218	347\\
217	346\\
217	345\\
216	344\\
215	343\\
214	343\\
213	342\\
212	341\\
211	340\\
210	339\\
209	338\\
208	337\\
207	336\\
206	336\\
205	335\\
204	335\\
203	334\\
202	334\\
201	334\\
200	334\\
199	334\\
198	334\\
197	334\\
196	334\\
195	334\\
194	333\\
193	333\\
192	332\\
191	332\\
190	332\\
189	332\\
188	332\\
187	332\\
186	332\\
185	332\\
184	332\\
183	332\\
182	333\\
181	333\\
180	333\\
179	334\\
178	334\\
177	334\\
176	334\\
175	334\\
174	334\\
173	334\\
172	335\\
171	336\\
170	336\\
169	336\\
168	336\\
167	336\\
166	337\\
165	337\\
164	337\\
163	337\\
162	337\\
161	336\\
160	336\\
159	336\\
158	336\\
157	335\\
156	335\\
155	334\\
154	334\\
153	333\\
152	333\\
151	333\\
150	332\\
149	332\\
148	332\\
147	331\\
146	331\\
145	330\\
144	330\\
143	329\\
142	329\\
141	328\\
140	327\\
139	327\\
138	326\\
137	326\\
136	325\\
135	324\\
134	323\\
133	322\\
132	321\\
131	320\\
130	319\\
129	318\\
128	317\\
127	316\\
126	315\\
125	314\\
124	313\\
123	312\\
122	311\\
121	310\\
120	309\\
119	309\\
118	309\\
118	308\\
117	307\\
117	306\\
116	305\\
116	304\\
115	303\\
115	302\\
115	301\\
115	300\\
114	299\\
114	298\\
114	297\\
114	296\\
113	295\\
113	294\\
113	293\\
112	292\\
112	291\\
112	290\\
111	289\\
111	288\\
111	287\\
111	286\\
111	285\\
111	284\\
111	283\\
111	282\\
110	281\\
110	280\\
110	279\\
110	278\\
110	277\\
110	276\\
110	275\\
110	274\\
110	273\\
110	272\\
110	271\\
110	270\\
110	269\\
110	268\\
110	267\\
110	266\\
110	265\\
110	264\\
110	263\\
110	262\\
110	261\\
110	260\\
110	259\\
111	258\\
111	257\\
111	256\\
111	255\\
111	254\\
111	253\\
111	252\\
112	251\\
112	250\\
113	249\\
113	248\\
114	247\\
114	246\\
114	245\\
114	244\\
115	243\\
115	242\\
116	241\\
117	240\\
118	239\\
118	238\\
119	237\\
119	236\\
120	235\\
121	234\\
122	233\\
123	232\\
123	231\\
124	230\\
125	230\\
126	229\\
127	228\\
127	227\\
128	226\\
129	225\\
130	225\\
131	224\\
132	224\\
133	223\\
134	222\\
135	221\\
136	220\\
137	220\\
138	220\\
139	219\\
140	218\\
141	218\\
142	217\\
143	216\\
144	215\\
145	214\\
146	214\\
147	214\\
148	213\\
149	212\\
150	212\\
151	211\\
152	210\\
153	210\\
154	210\\
155	210\\
156	210\\
157	210\\
158	210\\
159	210\\
160	210\\
161	209\\
162	209\\
163	208\\
164	208\\
165	208\\
166	208\\
167	208\\
168	208\\
169	208\\
170	208\\
171	208\\
172	208\\
173	208\\
174	208\\
175	208\\
176	208\\
177	207\\
178	207\\
179	207\\
180	207\\
181	207\\
182	207\\
183	207\\
184	207\\
185	207\\
186	207\\
187	208\\
188	208\\
189	208\\
190	208\\
191	208\\
192	209\\
193	209\\
194	209\\
195	210\\
196	211\\
197	212\\
198	212\\
199	212\\
200	212\\
201	213\\
202	213\\
203	214\\
204	214\\
205	214\\
206	215\\
207	216\\
208	216\\
209	217\\
210	218\\
211	218\\
212	219\\
213	220\\
214	220\\
215	221\\
216	221\\
217	221\\
217	222\\
217	223\\
218	224\\
219	224\\
220	224\\
221	225\\
222	225\\
223	225\\
224	226\\
225	227\\
226	228\\
227	229\\
228	230\\
229	231\\
229	232\\
230	233\\
231	234\\
232	235\\
232	236\\
233	237\\
233	238\\
234	239\\
235	240\\
235	241\\
236	242\\
237	243\\
238	244\\
238	245\\
239	246\\
239	247\\
239	248\\
240	249\\
240	250\\
241	251\\
242	251\\
243	251\\
244	252\\
245	253\\
246	253\\
247	254\\
248	254\\
249	255\\
250	256\\
251	257\\
252	257\\
253	258\\
254	258\\
255	259\\
256	259\\
257	260\\
258	261\\
259	261\\
260	261\\
261	261\\
262	262\\
263	262\\
264	262\\
265	262\\
266	262\\
267	262\\
268	262\\
269	262\\
270	262\\
271	262\\
272	262\\
273	262\\
274	262\\
275	262\\
276	262\\
277	262\\
278	262\\
279	262\\
280	262\\
281	262\\
282	262\\
283	261\\
284	261\\
285	261\\
286	260\\
287	259\\
288	259\\
289	259\\
290	258\\
291	257\\
292	256\\
293	255\\
294	254\\
295	254\\
295	253\\
296	253\\
296	252\\
297	252\\
297	251\\
298	251\\
298	250\\
299	249\\
299	248\\
300	248\\
300	247\\
300	246\\
301	246\\
301	245\\
302	245\\
303	244\\
304	244\\
304	243\\
305	242\\
305	241\\
306	240\\
307	240\\
308	240\\
309	240\\
309	239\\
310	238\\
311	237\\
311	236\\
312	236\\
313	236\\
313	235\\
314	235\\
315	235\\
316	235\\
316	234\\
317	234\\
318	234\\
318	233\\
319	233\\
319	232\\
320	232\\
321	232\\
322	232\\
323	232\\
324	232\\
325	232\\
326	232\\
326	233\\
327	233\\
327	234\\
328	234\\
328	235\\
330	235\\
331	235\\
332	235\\
332	236\\
333	236\\
334	236\\
334	237\\
335	237\\
336	237\\
336	238\\
337	238\\
338	238\\
339	238\\
340	238\\
341	239\\
342	239\\
343	240\\
344	241\\
344	242\\
345	243\\
346	244\\
347	245\\
348	246\\
348	247\\
349	248\\
350	249\\
351	250\\
352	250\\
353	251\\
353	252\\
354	253\\
355	254\\
355	255\\
356	256\\
357	257\\
357	258\\
358	259\\
358	260\\
358	261\\
359	262\\
359	263\\
360	264\\
361	265\\
361	266\\
362	267\\
363	268\\
363	269\\
363	270\\
363	271\\
364	272\\
365	273\\
366	274\\
366	275\\
367	276\\
368	277\\
369	278\\
369	279\\
369	280\\
370	281\\
370	282\\
370	283\\
371	284\\
372	285\\
372	286\\
373	287\\
373	288\\
373	289\\
374	290\\
375	291\\
376	292\\
376	293\\
376	294\\
377	295\\
377	296\\
377	297\\
377	298\\
377	299\\
377	300\\
378	301\\
378	302\\
378	303\\
378	304\\
378	305\\
379	306\\
379	307\\
379	308\\
380	309\\
380	310\\
380	311\\
380	312\\
380	313\\
380	314\\
381	315\\
381	316\\
381	317\\
380	318\\
380	319\\
379	320\\
378	321\\
377	322\\
377	323\\
376	324\\
375	325\\
374	326\\
374	327\\
373	328\\
373	329\\
372	330\\
372	331\\
372	332\\
371	333\\
370	334\\
370	335\\
369	336\\
369	337\\
369	338\\
368	339\\
367	340\\
367	341\\
366	342\\
366	343\\
365	344\\
365	345\\
364	346\\
363	347\\
363	348\\
363	349\\
362	350\\
362	351\\
362	352\\
361	353\\
361	354\\
360	355\\
360	356\\
360	357\\
360	358\\
359	359\\
359	360\\
359	361\\
359	362\\
358	363\\
358	364\\
358	365\\
358	366\\
357	367\\
357	368\\
357	369\\
356	370\\
356	371\\
355	372\\
355	373\\
355	374\\
355	375\\
355	376\\
355	377\\
355	378\\
354	379\\
354	380\\
354	381\\
354	382\\
354	383\\
354	384\\
354	385\\
354	386\\
354	387\\
354	388\\
354	389\\
354	390\\
354	391\\
354	392\\
354	393\\
354	394\\
354	395\\
354	396\\
354	397\\
354	398\\
354	399\\
354	400\\
354	401\\
354	402\\
354	403\\
354	404\\
354	405\\
353	406\\
353	407\\
353	408\\
353	409\\
353	410\\
353	411\\
353	412\\
353	413\\
353	414\\
353	415\\
353	416\\
353	417\\
353	418\\
353	419\\
353	420\\
353	421\\
353	422\\
353	423\\
353	424\\
353	425\\
353	426\\
353	427\\
353	428\\
353	429\\
353	430\\
353	431\\
353	432\\
353	433\\
353	434\\
353	435\\
352	436\\
352	437\\
352	438\\
352	439\\
352	440\\
352	441\\
352	442\\
352	443\\
352	444\\
};

\end{axis}
\end{tikzpicture}%
\end{subfigure}
\begin{subfigure}[b]{0.23\textwidth}
\input{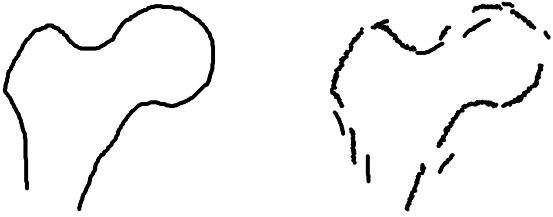}
\end{subfigure}
\caption{Comparison of an original contour (left) and a contour with artificial distortion for the outlier test (right). }\label{fig:outliers}
\end{figure}
\subsection{Test of performance}
We do not have available a ground truth of the registration and correspondence parameters. Therefore, to evaluate the accuracy of registration, we consider two different metrics, each with their own particular limitations. To test the robustness against the presence of outliers, we design a test in which we simulate typical errors that an automatic edge detector might introduce into the segmentation. Additionally, since DTW requires ordered input signals, we study the impact of an unknown topology with unordered inputs. Finally, we consider the total variance of the group-wise registration as a quality measure.

\begin{figure*}[t]
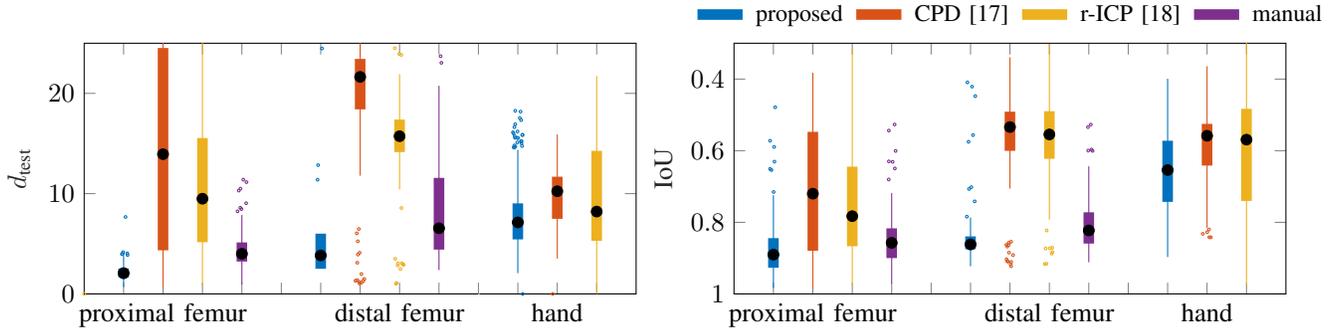

\input{boxplot_dtest}
\input{boxplot_IoU}
\caption{Boxplots comparing the registration results using our proposed strategy (blue), CPD \cite{Myronenko10} (red), r- (robust) ICP \cite{Bergstrom17} (yellow), and manually determined correspondences (purple). There are no manual correspondences available for the hand shapes. The left plot shows $d_{\text{test}}$, which is measured in pixels (in the fluoroscopic images the pixel size is $\approx 0.45$ mm, and in the hand images $\approx 0.5$ mm \cite{Yoruk06}). The right plot shows IoU. In Algorithms \ref{alg:alt1} and \ref{alg:alt2} we have employed $i_{\text{max}} = 100$ (maximum number of iterations) and $c_{\text{min}} = 10^{-4}\min_{m=1,\dots,M} ||\mathbf{x}_m||$ (tolerance stopping criterion).}\label{fig:BoxPlot}
\end{figure*}

\begin{figure}[t]
\input{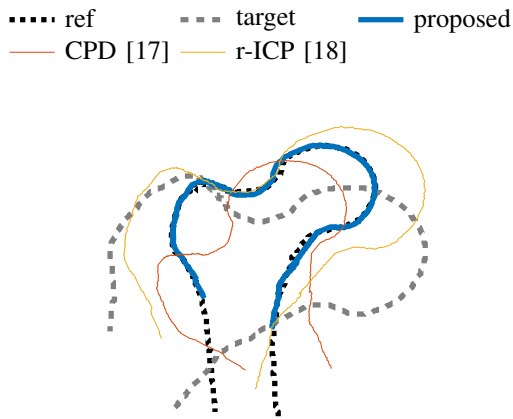}
\caption{Comparative example of the proposed registration of the contour of the proximal femur. The reference (ref) and target have different lengths, sizes, and original positions.}\label{fig:FHex}
\end{figure}
\subsubsection{Accuracy of pairwise registration} \label{sec:testdist}
Since there is no ground truth of the registration and correspondence parameters, we consider the following performance metrics. First, the sum of minimum distances between a reference vector $\mathbf{x}_{\text{ref}} \in \mathbb{C}^{N_r}$ and a target vector $\mathbf{x}_{\text{target}} \in \mathbb{C}^{N_t}$, that is
\begin{equation}
d_{\text{test}} = \frac{1}{N_t}{\sum_{n_t=1}^{N_t} \min_{n_r = 1, \dots, N_r}|x_{\text{target}}[n_t]-{x}_{\text{ref}}[n_r]|}.
\end{equation}
This distance can be interpreted as a modified Hausdorff distance \cite{Yoruk06}.
The interpretation of $d_{\text{test}}$ may be misleading, for instance, in the registration of the hand contours, where a short distance may be achieved if a finger in the target is aligned to an incorrect finger in the reference.
We therefore also evaluate a second metric: the intersection-over-union (IoU) of the areas that the contours cover after the registration \cite{Crum06}. However, this metric may also be misleading for the evaluation of incomplete contours, as is the case in our femur database, since a perfect registration may represent a small intersection area with respect to the union area. Unfortunately, there does not seem to be one single quantity that perfectly measures the accuracy of registration in this problem.
 
\subsubsection{Outliers}
Outliers are points that do not belong to the contour or have been distorted and disrupt the shape. The femur contours in our database are almost free of outliers because they were manually obtained. The competing strategies \cite{Myronenko10} and \cite{Bergstrom17} are designed to deal with outliers and noisy contours. In order to compare the robustness of our strategy to the competing techniques, we perform a test where we add outliers artificially to our database by emulating the distortion possibly induced by automatic segmentation. First, we add a noise component to the shape vector $\mathbf{x}_m$, i.e., $\mathbf{z}_{m}= \mathbf{x}_m+\boldsymbol{\gamma}$, with $\boldsymbol{\gamma} \sim \mathcal{CN}(0,\sigma^2_{n}\mathbf{I})$, to consider a small noisy deformation. Then, for each noisy shape vector, $\mathbf{z}_m$, we randomly contort 10 segments of random length $l_s$, in different sections of the contour, each starting at a random index $n_s$. Let us denote the points in these segments as $\mathbf{z}_{\text{outliers}}^{(s)} $, for $s = 1, \dots, 10$. We displace the points from the contour emulating the typical errors of an edge detector occurring when an edge from a neighboring structure is detected instead of the true contour. The resulting displaced and distorted segment is
\begin{equation}
 \mathbf{z}_{\text{outliers}}^{(s)} =  \mathbf{z}^{(s)} + \mathbf{1}\beta_s,
\end{equation}
where $\mathbf{z}^{(s)} = \big[z_m[n_s],\dots,z_m[n_s+l_s]\big]^T$ and $\beta_s \sim \mathcal{CN}(0,\sigma^2_{t})$. We choose $\sigma^2_{t}$, $\sigma^2_{n}$ and the range of $l_s$ by visual inspection, such that the resulting contours look realistic. We show an example of a contour from the proximal femur database affected by such displacements and distortions in Fig. \ref{fig:outliers}, where $\sigma_{t} = 12$ pixels, $\sigma_{n}=1$ pixel, and $l_s$ is a uniform random value between $1$ and $10$ percent of the total length of the contour.

\subsubsection{Unknown order}
DTW requires that the order of the entries of each vector follow the contour, with the first entry corresponding to the start point and the last entry corresponding to the end point. 
In our database this is the case because the contours were manually determined. However, when the contours are automatically extracted, e.g. with an edge detector, this order may be unknown. The inputs to the competing techniques are point clouds, with no given topology. In order to determine to what extent the performance of our technique depends on an a priori known order, we propose the following test: We randomly shuffle the points in the shape vectors so that the order is unknown and add a preprocessing step to estimate the order before applying our registration technique. Start and end points are determined heuristically. Since we assume that the target and reference follow the same shape model, we have not contemplated mirroring. We may assume w.l.o.g. to follow the contour in an anti-clockwise direction, and the start and end points are assigned based on their proximity to the image border. Then, we use alpha shapes \cite{Edelsbrunner10} to calculate an approximate order and reorder the points accordingly.

\subsubsection{Performance of the group-wise registration}
Following the properties of a good shape model described in \cite{Davies02} and \cite{Davies08}, we evaluate the total variance of the group-wise result as a quality metric. We compare our proposed extended generalized Procrustes alignment to a regular generalized Procrustes alignment, which is based on landmarks whose correspondence was manually determined.
\section{Results and discussion}

In this section, we study the accuracy of the registration of a pair of shape vectors in comparison with the competing strategies described in Section \ref{sec:CompStr}. We also compare our group-wise approach with a generalized Procrustes analysis based on manual correspondences, since this is the typical solution for designing statistical shape models. 

\subsection{Procrustes registration of two shape vectors} 
We first discuss the accuracy of the pair-wise registration. We evaluate the distance of registering each shape vector to the longest vector in each data set. We measured the described distance $d_{\text{test}}$ and the IoU. We show the resulting boxplots in Fig. \ref{fig:BoxPlot}. We considered the femoral shapes from our database (proximal and distal) as well as the hand shapes. Our strategy performs best median results among all competing techniques, for all three data sets, in terms of both metrics $d_{\text{test}}$ and IoU. It also has the smallest variability in performance as measured by these metrics. Our technique even outperforms the manually determined correspondences (where available). We show one example of the registration results in Fig. \ref{fig:FHex}, where our strategy performed a much more accurate registration.

\begin{figure}[t]
\centering
\input{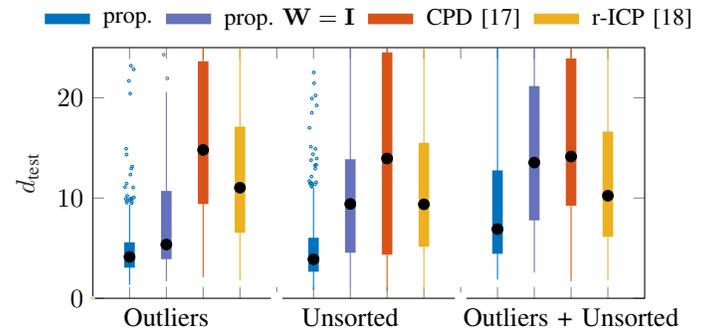}
\caption{Boxplot of error metric $d_{\text{test}}$ in pixels (pixel size $\approx 0.45$ mm) for registration with outliers and/or without prior ordering, for the proximal femur database. We considered our proposed strategy (prop., blue) as well as an unweighted version of our proposed strategy (prop. $\mathbf{W}=\mathbf{I}$, violet). }

\label{fig:outliers_boxplot}
\end{figure}

\begin{figure}[t]
\input{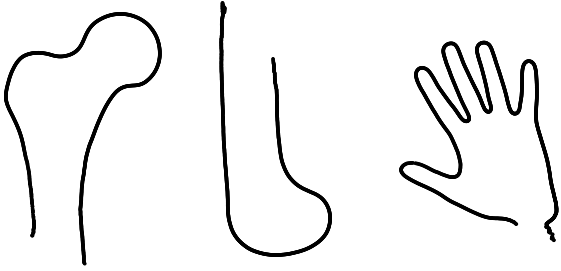}

\caption{Estimated mean shape (proximal femur, distal femur, hand) obtained with proposed group-wise solution as described in Algorithm \ref{alg:alt2}.}\label{fig:GroupWise}
\end{figure}

\begin{figure}[t]
%
%
\begin{tikzpicture}

\begin{axis}[%
width=0.33\figurewidth,
at={(0in,0in)},
scale only axis,
point meta min=0,
point meta max=20,
axis on top,
x dir=reverse,
xmin=-28.9570552147239,
xmax=88.9570552147239,
y dir=reverse,
ymin=0.5,
ymax=93.5,
axis line style={draw=none},
ticks=none,
legend style={legend cell align=left, align=left, draw=white!15!black}
]
\addplot [forget plot] graphics [xmin=0.5, xmax=59.5, ymin=0.5, ymax=93.5] {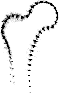};
\end{axis}
\end{tikzpicture}%
%
%
\begin{tikzpicture}
\begin{axis}[%
width=0.33\figurewidth,
at={(0in,0in)},
scale only axis,
point meta min=0,
point meta max=20,
axis on top,
x dir=reverse,
xmin=-28.9570552147239,
xmax=88.9570552147239,
y dir=reverse,
ymin=0.5,
ymax=93.5,
axis line style={draw=none},
ticks=none,
legend style={legend cell align=left, align=left, draw=white!15!black}
]
\addplot [forget plot] graphics [xmin=0.5, xmax=59.5, ymin=0.5, ymax=93.5] {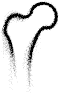};
\end{axis}
\end{tikzpicture}%
%
%
\begin{tikzpicture}

\begin{axis}[%
width=0.33\figurewidth,
at={(0in,0in)},
scale only axis,
point meta min=0,
point meta max=20,
axis on top,
xmin=-37.3323108384458,
xmax=98.3323108384458,
y dir=reverse,
ymin=0.5,
ymax=107.5,
axis line style={draw=none},
ticks=none,
legend style={legend cell align=left, align=left, draw=white!15!black}
]
\addplot [forget plot] graphics [xmin=0.5, xmax=60.5, ymin=0.5, ymax=107.5] {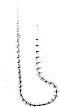};
\end{axis}
\end{tikzpicture}%
%
%
\begin{tikzpicture}
\begin{axis}[%
width=0.33\figurewidth,
at={(0in,0in)},
scale only axis,
point meta min=0,
point meta max=20,
axis on top,
x dir=reverse,
xmin=-37.3323108384458,
xmax=98.3323108384458,
y dir=reverse,
ymin=0.5,
ymax=107.5,
axis line style={draw=none},
ticks=none,
legend style={legend cell align=left, align=left, draw=white!15!black}
]
\addplot [forget plot] graphics [xmin=0.5, xmax=60.5, ymin=0.5, ymax=107.5] {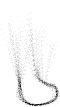};
\end{axis}
\end{tikzpicture}%

\caption{40 equidistant points from proximal and distal femoral contours after applying the proposed Algorithm \ref{alg:alt2} (left images), and a typical generalized Procrustes analysis that is based on manual landmarks (right images).} 
\label{fig:compare_compact}

\end{figure}
\subsection{Outliers and unknown order} 
To determine how well the techniques handle outliers and possibly unknown order of points, we perform two further registration tests for the proximal femur contours. 

\subsubsection{Test with outliers}
In Fig. \ref{fig:outliers_boxplot} (labeled as ``Outlier") we show a boxplot of the registration results with the outlier model discussed in Section \ref{sec:DetW} in terms of $d_{\text{test}}$. We observe that our strategy, even though it experiences some loss of accuracy with respect to the outlier-free result, still outperforms the competing strategies. Moreover, if the proposed weighted least-squares minimization in \eqref{WLS} is substituted with an ordinary least-squares approach, i.e., $\mathbf{W}=\mathbf{I}$, there is some loss in performance. Hence, the weighted least-squares registration adds robustness.

\subsubsection{Test with unknown order}
In Fig. \ref{fig:outliers_boxplot} (labeled as ``Unsorted") we present the results of a test in which the points in the vector were shuffled. As expected, this has some effect on our technique but not the competitors since those do not use a prior point ordering. However, our strategy still outperforms the competition, and the weights in \eqref{WLS} enhance the accuracy.
\subsubsection{Test with outliers and unknown order}
In Fig. \ref{fig:outliers_boxplot} (labeled as ``Outliers + Unsorted") we show the results of the combined effects of outliers and unsorted vectors. Our proposed registration still outperforms all competing techniques.
\subsection{Group-wise registration}

We performed a group-wise registration of each of our three shape data sets. 
We show the results of the obtained mean shape in Fig. \ref{fig:GroupWise}, which look as expected for each dataset. The estimate is worst at the extremes of the femur contours due to a smaller number of samples in these regions. For the hands, the wrist experiences more outliers and variability \cite{Yoruk06}.

We compare the result of our group-wise registration to a registration based on manual correspondences. We performed this study with the femoral data sets, where manual landmarks are available. We show the qualitative results in Fig. \ref{fig:compare_compact}. In order to improve the visualization, we show only 40 points, equidistantly chosen on the registered contours. We see that the proposed registration provides a more compact representation. As a quantitative metric we also evaluate the total variance of the points, which is the trace of the sample covariance matrices of the registered vectors. This metric is used to quantify the quality of the correspondences in \cite{Davies08}. The ratio between the total variances of the proposed and the manual models is $0.05$ for the femoral head, and $0.1$ for the condyle. This means that the total variance of the proposed registration is one order of magnitude smaller, and hence more compact and better registered \cite{Davies08}.

\section{Conclusions}
An accurate registration is fundamental in segmentation and retrieval applications that use a shape model, as well as for learning the models.  In order to avoid the manual selection of landmarks in statistical shape model training, we have investigated the problem of registration when there are not correspondence references. Our technique saves annotation time and does not require manually placed landmarks. We have solved simultaneously the correspondence and the Procrustes registration
problem with an alternating optimization approach. The proposed correspondences are based on DTW, and a probabilistic
alignment that adds robustness. DTW preserves the order of points on a contour, providing higher accuracy and preserving efficiency, which is the main difference with respect to the strategies based on the ICP algorithm. 
Our strategy outperformed competing approaches with similar computation time when tested on three different anatomical contours, even in the presence of outliers and with unordered points.
We have also proposed a group-wise solution, which provides a training set that is accurately registered, and hence more compact and meaningful shape models can be learned.

However there are still some issues that have to be addressed, which we enumerate in the following. The group-wise registration is dependant on the sample that is initially chosen as the mean, which may lead to miss deformation details in the first and second order statistics of the obtained shape model. A more sophisticated strategy to deal with the multiple-to-one correspondences in the group-wise registration needs to be designed in order to account for these features. Furthermore, our approach has not considered the presence of important occlusions in the training samples. From the observation of our outlier test we can infer that, if these occlusions are not very significant, our current algorithm should be robust enough to deal with them. However, a more thorough test would be require to verify this, and possibly an additional enhancement of the correspondence strategy could be added to deal with occlusions in future work. Finally, a limitation of our strategy, especially for clinical application, is the two-dimensional design. In order to deal with three-dimensional anatomical models, that are present in medical imaging with stronger importance that the two-dimensional ones, the proposed approach needs to be extended. Our future work is focused on improving these limitations. Our efforts are especially the three-dimensional extension of the approach.

Nevertheless, to the best of our knowledge, the use of DTW in a Procrustes registration for statistical shape model analysis has not yet been explored. In this contribution we show that the registration accuracy is improved with a DTW-correspondence, and that the tedious task of manual labelling of the training landmarks can be avoided to train the models. We believe that DTW is a promising tool to enhance registration algorithms.

\bibliographystyle{IEEEbib}
\bibliography{references}

\end{document}